\documentclass{article}

 \usepackage[preprint]{neurips_2026}
\usepackage{graphicx}  
\usepackage[utf8]{inputenc} 
\usepackage[T1]{fontenc}    
\usepackage{hyperref}       
\usepackage{url}            
\usepackage{booktabs}       
\usepackage{amsfonts}       
\usepackage{nicefrac}       
\usepackage{amsmath}
\usepackage{amssymb}
\usepackage{amsthm}
\usepackage{microtype}      
\usepackage{xcolor}         
\usepackage[table]{xcolor}
\usepackage{algorithm}
\usepackage{algorithmic}
\usepackage{multirow}
\usepackage{subcaption}

\usepackage[most]{tcolorbox}

\definecolor{mydarkblue}{rgb}{0,0.08,0.45}

\setcitestyle{authoryear,round,citesep={;},aysep={,},yysep={;}}

\makeatletter
\renewcommand{\NAT@open}{\begingroup\color{mydarkblue}(}
\renewcommand{\NAT@close}{)\endgroup}
\makeatother

\definecolor{PromptGreen}{HTML}{F1FAF1}
\definecolor{PromptGreenFrame}{HTML}{8BC78B}
\definecolor{PromptGreenTitle}{HTML}{2F6F3A}
\definecolor{MMPOGray}{HTML}{F2F2F2}
\newtcolorbox{promptbox}{
    enhanced,
    width=\linewidth,
    colback=PromptGreen,
    colframe=PromptGreenFrame,
    boxrule=0.6pt,
    arc=1.5mm,
    left=8pt,
    right=8pt,
    top=7pt,
    bottom=7pt,
    before skip=0.8em,
    after skip=0.8em,
}
\title{Meta-Cognitive Memory Policy Optimization for Long-Horizon LLM Agents}

\author{%
  \textbf{Ziyan Liu\textsuperscript{1 *}}
  \quad
  \textbf{Zhezheng Hao\textsuperscript{2 *}}
  \quad
  \textbf{Yeqiu Chen\textsuperscript{1 *}}
  \quad
  \textbf{Hong Wang\textsuperscript{1}}
  \quad
  \textbf{Jingren Hou\textsuperscript{1}}
  \\[0.4em]
  \textbf{Ruiyi Ding\textsuperscript{1}}
  \quad
  \textbf{Yongkang Yang\textsuperscript{1}}
  \quad
  \textbf{Wence Ji\textsuperscript{3}}
  \quad
  \textbf{Wei Xia\textsuperscript{3 \textdagger}}
  \quad
  \textbf{Feng Liu\textsuperscript{3}}
  \\[0.6em]
  \normalfont
  \textsuperscript{1}University of Science and Technology of China
  \quad
  \textsuperscript{2}Zhejiang University
  \quad
  \textsuperscript{3}Tencent
  \\[0.3em]
  {\small
  \textsuperscript{*}Equal contribution.
  \quad
  \textsuperscript{\textdagger}Corresponding author: \texttt{xwellxia@tencent.com}.
  }
}

\begin{document}

\maketitle

\begin{abstract}
Memory-augmented LLM agents tackle complex long-horizon tasks by recursively summarizing interaction trajectories into compact memory. However, existing approaches typically train these memory policies using outcome-based reinforcement learning, failing to localize where intermediate memory quality degrades. As interactions unfold, ambiguous recursive summaries progressively discard task-relevant information and introduce semantic noise. This exacerbates belief deviation, obscuring the agent’s estimate of the latent task state and ultimately derailing long-horizon reasoning. We therefore argue that memory optimization should focus not merely on trajectory-level success, but on the clarity of the belief induced by intermediate summaries. To this end, we introduce Belief Entropy, a self-supervised proxy that probes how uncertain the model remains about the latent task state given its current memory. Based on this proxy, we propose Metacognitive Memory Policy Optimization (MMPO). Instead of relying only on sparse outcome-based signals, MMPO provides fine-grained, memory-specific supervision via explicitly penalizing summaries that induce high epistemic uncertainty. 
Experiments show that MMPO consistently outperforms existing methods on diverse long-horizon tasks, maintaining 97.1\% performance even when scaled to 1.75M-token contexts.
\end{abstract}

\section{Introduction}
Solving complex problems in long-horizon environments with reliable internal memory forms a cornerstone of human intelligence, which is also an important component in building Artificial General Intelligence (AGI). Recently, Large Language Models (LLMs)~\citep{Survey-LLMs, GPT4, Llama3, Deepseek-V3} have demonstrated remarkable reasoning capabilities. However, during extended long-horizon interactions, they are often bottlenecked by limited context windows and the "lost-in-the-middle" phenomenon~\citep{lost-in-the-middle, long-context-survey}. To address these limitations, memory-augmented agents have emerged as a prominent paradigm~\citep{du2026memory, MemGPT, MemOS, Mem0}, which recursively summarize past interaction trajectories into compact memory. This compressed memory enables the agent to continuously reason and execute tasks within a consistently bounded context window.

Despite this advantage, recursive summarization inherently accumulates semantic noise introduced by LLMs, which can induce cascading hallucinations and ultimately degrade long-horizon collapse~\citep{zhang2023siren, ji2023survey,sheng2026when}. To improve internal memory management, recent work typically adopts Reinforcement Learning with Verifiable Rewards (RLVR) to train summary policies based on final outcome success or failure~\citep{RL4LRM, Deepseek-R1, MemAlpha, MemAgent}.
However, training with such sparse rewards introduce a severe credit assignment problem: it fails to localize intermediate memory degradation and provides no explicit supervision to suppress noise accumulation during recursive summarization.
Consequently, the agent remains prone to accumulating noisy or irrelevant information in memory, leading to \textit{memory explosion} and performance decay as the interactions unfold~\citep{sheng2026when}.
This limitation stems from the lack of a principled criterion for intermediate summary optimization.

\begin{figure}[t]
    \centering
    \includegraphics[width=\textwidth]{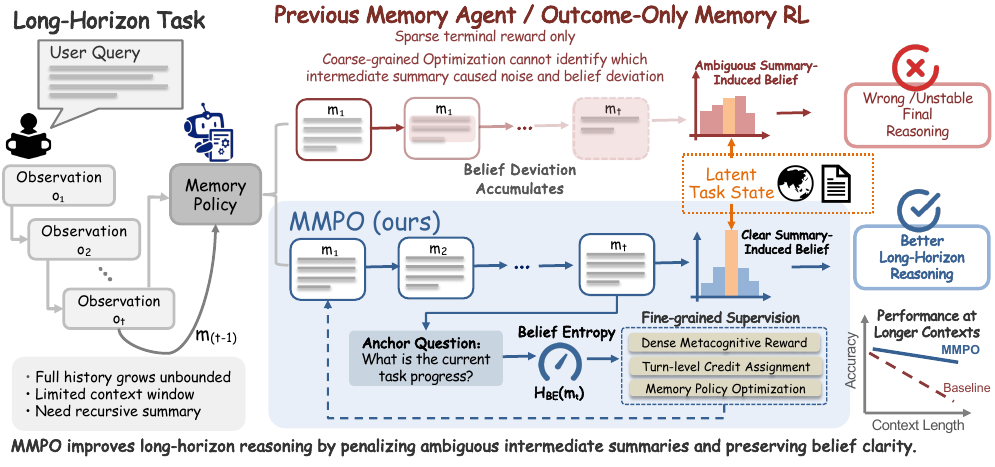}
    \vspace{-2em}
\caption{\textbf{Overview of MMPO.} \textbf{(Top)} Existing outcome-based memory policies suffer from sparse credit assignment, failing to prevent ambiguous summaries from accumulating \textit{belief deviation}. \textbf{(Bottom)} MMPO introduces an anchor-question-based \textit{Belief Entropy} to provide dense, memory-specific supervision. This fine-grained penalty for epistemic uncertainty preserves clearer summary-induced beliefs and improves long-context reasoning.}
    \label{fig:hero}
    \vspace{-1.5em}
\end{figure}


To address this limitation, we first analyze what accumulated summary noise disrupts in the agent’s decision process. We formulate multi-turn agentic tasks as Partially Observable Markov Decision Processes (POMDPs), where hidden task states require the agent to act according to a \textbf{belief state}~\citep{astrom1965optimal, kaelbling1998planning}—an internal probabilistic estimate derived from the interaction history. Under this formulation, recent work~\citep{zoureducing} has attributed long-horizon reasoning collapse to \textit{belief deviation}, i.e., the progressive drift between the agent’s internal belief and the underlying latent task state as interactions extend. In summary-based workflows, textual memory replaces the full interaction history as the agent’s decision context, thereby inducing the belief that guides subsequent reasoning and actions.Therefore, semantic noise accumulated through recursive summarization manifests as deviation in this \textit{summary-induced} belief. This makes \textit{belief preservation} the principled criterion for intermediate memory optimization: a reliable intermediate summary should maintain an accurate and stable estimate of the current underlying task state.

This perspective suggests that final task outcomes are insufficient for memory optimization: reliable long-horizon memory requires fine-grained supervision on the clarity of summary-induced belief. However, directly measuring this belief uncertainty is infeasible in open-ended LLM settings, since the latent task state is not observable. To bridge this gap, we adopt a metacognitive probe inspired by cognitive science~\citep{flavell1979metacognition, nelson1990metamemory} to estimate the model’s intrinsic uncertainty about the task state from the current memory. Instantiated with a dedicated \textit{anchor question}, this probe yields \textbf{Belief Entropy}, a self-supervised signal that measures response uncertainty as a proxy for summary-induced belief clarity. We empirically validate Belief Entropy as a reliable intermediate signal for memory optimization. Based on this signal, we propose \textit{Metacognitive Memory Policy Optimization} (MMPO), which augments outcome-based RL with Belief Entropy rewards at intermediate memory states, providing dense, memory-specific supervision beyond sparse final outcomes. Extensive experiments demonstrate that MMPO improves the performance of long-horizon memory agents. We further show that MMPO effectively reduces belief uncertainty and improves long-horizon reasoning stability.

\vspace{-1em}
\section{Belief Entropy}
\vspace{-0.75em}
\label{sec:theory}
\subsection{POMDP Formulation}
\label{sec:pomdp_formulation}
\vspace{-0.5em}
\begin{figure}[h]
    \centering
    \includegraphics[width=\textwidth]{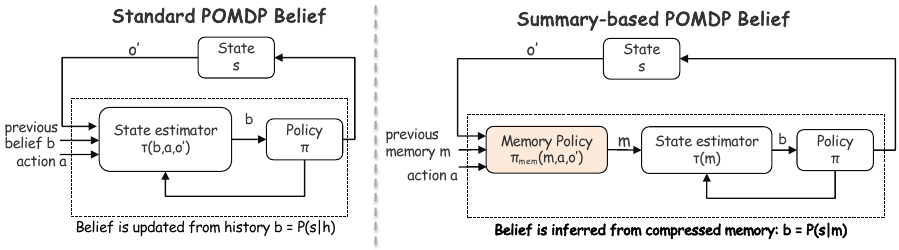}
    \caption{\textbf{Belief-state under standard and summary-based POMDPs.} 
    (a) In standard POMDPs, the belief $b=P(s\mid h)$ is updated from the full interaction history. 
    (b) In summary-based POMDPs, the memory policy compresses the history into a summary $m$, inducing a belief $b=P(s\mid m)$ from the compressed representation. }
    \label{fig:belief}
    \label{fig:belief}
\end{figure}

We model long-horizon reasoning as a Partially Observable Markov Decision Process (POMDP)~\citep{astrom1965optimal, smallwood1973optimal, kaelbling1998planning}: $\mathcal{M} = \langle \mathcal{S}, \mathcal{A}, \Omega, \mathcal{T}, \mathcal{O}, \mathcal{R}, \gamma \rangle$. At each step $t$, the agent observes $o_t \in \Omega$ (e.g., a retrieved document snippet) while the true task state $s_t \in \mathcal{S}$ remains hidden. It then executes action $a_t \in \mathcal{A}$; the environment transitions via $\mathcal{T}(s_{t+1} | s_t, a_t)$ and emits the next observation via $\mathcal{O}(o_{t+1} | s_{t+1}, a_t)$.

\paragraph{The Belief State.} Since $s_t$ is unobservable, the optimal policy $\pi^*(a_t | b_t)$ conditions on the \textit{belief state} $b_t \in \Delta(\mathcal{S})$, a distribution over hidden states. The belief summarizes the full interaction history $h_t = \{o_{\leq t}, a_{<t}\}$ as the posterior $b_t(s) \equiv P(s_t = s | h_t)$, updated recursively by the Bayesian filter:
$b_t(s') = \eta \cdot \mathcal{O}(o_t | s', a_{t-1}) \sum_{s \in \mathcal{S}} \mathcal{T}(s' | s, a_{t-1}) b_{t-1}(s)$, where $\eta = 1 / P(o_t | b_{t-1}, a_{t-1})$ normalizes. As a sufficient statistic of $h_t$, the belief provides all information needed for optimal decision-making.

\subsection{Belief Preservation for Memory Optimization}
\label{sec:memory_objective}
For long-horizon LLM agents, since the interaction history grows monotonically and suffers from ``lost-in-the-middle'' degradation~\citep{lost-in-the-middle, long-context-survey}, conditioning on the full history $h_t$ is impractical.
Such long-horizon partial observability induces \textit{belief deviation}, where the agent's internal state estimate to drift over extended interactions~\citep{zoureducing}.
To maintain finite context, memory-augmented agents use recursive summarization: at turn $t$, the memory policy updates a bounded textual summary $m_t = \pi_{\text{mem}}(m_{t-1}, a_{t-1}, o_t)$, and the action policy selects actions conditioned on this memory, $a_t \sim \pi_{\text{act}}(\cdot \mid m_t)$.
This limits long-horizon task execution within a fixed context budget.

\paragraph{Summary-Induced Belief.}
Because $m_t$ is produced by compressing the full interaction history $h_t$, summarization induces the Markov chain $s_t \rightarrow h_t \rightarrow m_t$. Since the action policy conditions only on $m_t$, the agent's belief is induced by this compressed summary:
\begin{equation}
b^M_t(s) \triangleq P(s_t \mid m_t).
\end{equation}
We defer a detailed derivation of the summary-induced belief to Appendix~\ref{app:summary_belief}.

\paragraph{Belief-Preservation Objective.}
As discussed above, belief deviation is a key source of long-horizon instability. Therefore, intermediate memory optimization should preserve the summary-induced belief by keeping the latent task state predictable from the summary $m_t$. This gives the belief-preservation objective:
\begin{equation}
    \max_{\pi_{\mathrm{mem}}} \;
    \mathbb{E}_{s_t,m_t}
    \big[
        \log P(s_t \mid m_t)
    \big].
\end{equation}
Therefore, maximizing the expected log-likelihood of the latent state amounts to minimizing the conditional entropy $H(s_t\mid m_t)$:
\begin{equation}
    \label{eq:ideal_objective}
    \arg\max_{\pi_{\mathrm{mem}}}
    \mathbb{E}_{s_t,m_t}
    \big[
        \log P(s_t \mid m_t)
    \big]
    \Longleftrightarrow
    \arg\min_{\pi_{\mathrm{mem}}}
    H(s_t \mid m_t).
\end{equation}
Using the identity $I(s_t; m_t) = H(s_t) - H(s_t \mid m_t)$, where $H(s_t)$ does not depend on the summary representation $m_t$, minimizing $H(s_t \mid m_t)$ is equivalent to maximizing the mutual information between the summary and the latent task state.
Thus, the memory objective is to make $m_t$ an informative representation of the underlying task state, thereby preserving a reliable belief.

\subsection{Belief Entropy as a Practical Proxy}
\label{sec:belief_entropy}
Directly optimizing Eq.~(\ref{eq:ideal_objective}) is intractable since the latent state $s_t$ is not directly observable. To bridge this gap, we use a metacognitive probe inspired by cognitive science~\citep{flavell1979metacognition, nelson1990metamemory, hart1965feeling} to estimate the model's intrinsic uncertainty about the task state from the current memory. We instantiate this probe with a task-state anchor question $q$ and define \textbf{Belief Entropy} as the uncertainty of the model's response to this anchor question $q$:
\begin{equation}
    \mathcal{H}_{\text{BE}}(m_t) \triangleq H(y \mid m_t, q), 
    \quad y \sim \pi_{\text{LLM}}(\cdot \mid m_t, q),
    \label{eq:belief_entropy}
\end{equation}
where $y$ denotes the model's response to the anchor question. Intuitively, a clear memory should induce a concentrated response distribution, while an ambiguous or incomplete memory should lead to higher response uncertainty. 

Following recent studies in entropy-based RLVR~\citep{hao2025rethinking, shen2026entropycontrol}, we estimate Belief Entropy by the mean of token-level predictive entropy.
Let $y^\ast$ be the greedy response to the anchor question, and let $\mathcal{V}_\ell$ denote the token set used for entropy computation at step $\ell$ (e.g., the full vocabulary or a compact top-$K$/top-$p$ candidate set with renormalized probabilities). We compute
\begin{equation}
    \widehat{\mathcal{H}}_{\mathrm{BE}}(m_t)
    =
    \frac{1}{|y^\ast|}
    \sum_{\ell=1}^{|y^\ast|}
    \left[
    -\sum_{v\in\mathcal{V}_\ell}
    \tilde{\pi}_{\mathrm{LLM}}(v \mid m_t,q,y^\ast_{<\ell})
    \log
    \tilde{\pi}_{\mathrm{LLM}}(v \mid m_t,q,y^\ast_{<\ell})
    \right],
    \label{eq:belief_entropy_token}
\end{equation}
Unless otherwise specified, all experimental uses of $\mathcal{H}_{\mathrm{BE}}$ refer to the empirical estimator $\widehat{\mathcal{H}}_{\mathrm{BE}}$.

\paragraph{The Design of Anchor Question.}
The anchor question is designed to convert unobservable belief uncertainty into observable response uncertainty. It should satisfy two criteria: it should condition explicitly on the current memory $m_t$, and it should probe task-state uncertainty rather than generic model confidence. We therefore use a dual-probe question: the progress component probes the agent's current task-state estimate, while the information-gap component probes residual uncertainty.

In practice, we adapt the following question to achieve such dual-probe: 

\begin{promptbox}
\centering
\textit{``Based on current memory, what is current task progress and what information is still needed?''}
\end{promptbox}

This design is motivated by the chain-rule 
$
\mathcal{H}_{\mathrm{BE}}(m_t)
=
H(y \mid m_t, q)
=
H(y \mid m_t, q, s_t)
+
I(y; s_t \mid m_t, q),
$
where the first term captures state-conditioned response uncertainty and the second term reflects residual state uncertainty exposed through the anchor response. The progress query probes the current state estimate, while the gap query probes unresolved uncertainty. Appendix~\ref{app:be_proof} further justifies the connection between Belief Entropy and $H(s_t \mid m_t)$.

\paragraph{Empirical Validation.}
We empirically validate $\mathcal{H}_{\text{BE}}$ on the MemAgent~\citep{MemAgent} with RULER-HotpotQA setting using Qwen2.5-7B. The analysis examines whether Belief Entropy behaves as a meaningful intermediate memory-quality signal before being used for policy optimization.

\begin{figure}[h]
\centering
\begin{subfigure}[t]{0.32\textwidth}
\centering
\includegraphics[width=\textwidth]{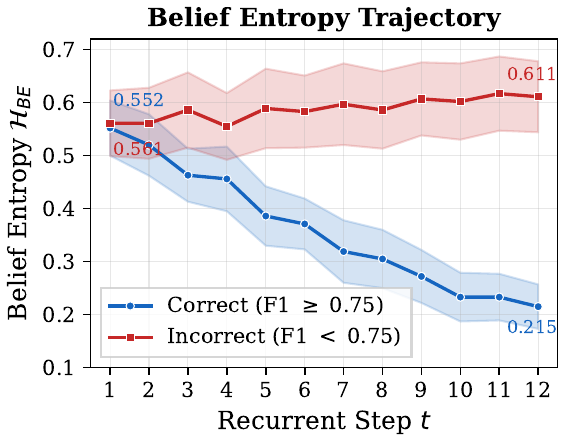}
\caption{BE trajectory trend.}
\label{fig:finding1}
\end{subfigure}
\hfill
\begin{subfigure}[t]{0.32\textwidth}
\centering
\includegraphics[width=\textwidth]{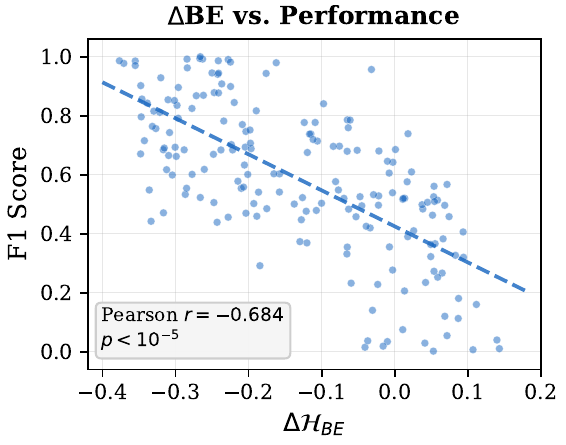}
\caption{$\Delta\mathcal{H}_{\text{BE}}$ vs.\ accuracy.}
\label{fig:finding2}
\end{subfigure}
\hfill
\begin{subfigure}[t]{0.32\textwidth}
\centering
\includegraphics[width=\textwidth]{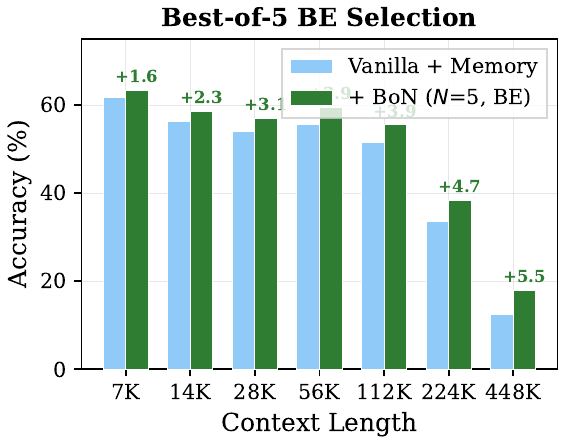}
\caption{Best-of-5 BE selection.}
\label{fig:finding3}
\end{subfigure}
\caption{Empirical validation of Belief Entropy. (a)~Successful trajectories show decreasing $\mathcal{H}_{\mathrm{BE}}$, while failed ones generally stagnate or increase. (b)~Entropy reduction correlates with task accuracy. (c)~Test-time Best-of-$N$ selection by $\mathcal{H}_{\mathrm{BE}}$ improves performance.}
\label{fig:findings}
\end{figure}

\textbf{Finding 1 (Trajectory Dynamics).}
Successful trajectories exhibit a consistent decrease in $\mathcal{H}_{\text{BE}}$ as relevant evidence is accumulated, whereas failed trajectories show increasing entropy (Figure~\ref{fig:findings}a). This suggests that lower Belief Entropy corresponds to a clearer summary-induced task state.

\textbf{Finding 2 (Outcome Correlation).}
The total entropy reduction $\Delta \mathcal{H}_{\text{BE}}$ is strongly correlated with final task accuracy (Pearson $r=-0.684$; Figure~\ref{fig:findings}b), indicating that stronger entropy reduction is empirically associated with better task performance.

\textbf{Finding 3 (Inference-Time Selection).}
Without any training, selecting the lowest-entropy trajectory among $N=5$ candidates improves accuracy over Vanilla+Memory (Figure~\ref{fig:findings}c). This demonstrates that $\mathcal{H}_{\text{BE}}$ provides an actionable memory-quality signal independent of the training objective.

\section{Metacognitive Memory Policy Optimization (MMPO)}
\label{sec:method}

Equipped with Belief Entropy as a tractable proxy for memory quality, we propose \textbf{Metacognitive Memory Policy Optimization (MMPO)}. MMPO addresses the credit assignment challenge in long-horizon reasoning by injecting dense process supervision at the level of intermediate summaries. We formalize this using a group-relative paradigm that evaluates sub-trajectories based on both their intermediate belief quality and their contribution to the final outcome. The complete training pipeline proceeds through three stages: trajectory sampling, Belief Entropy computation, and policy optimization.

\begin{figure}[h]
    \centering
    \includegraphics[width=\textwidth]{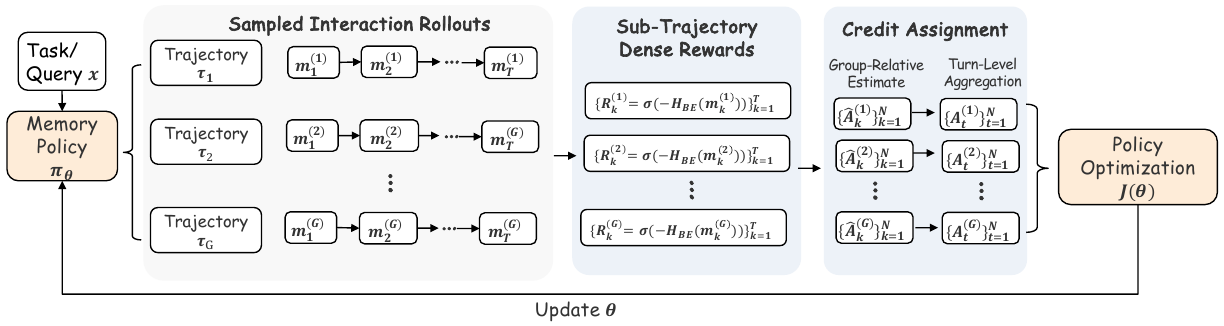}
    \caption{Overview of the MMPO training pipeline. \textbf{Stage~1:} The memory policy $\pi_\theta$ samples $G$ trajectories per task. \textbf{Stage~2:} Each trajectory is decomposed into sub-trajectories $\tau_{\leq 1}, \ldots, \tau_{\leq T}$, and Belief Entropy $\mathcal{H}_{\text{BE}}(m_k)$ is computed at every turn to produce dense per-step rewards $R_k$. \textbf{Stage~3:} Sub-trajectory rewards are normalized via GRPO and aggregated into future-aware turn-level advantages for policy optimization.}
    \label{fig:framework}
\end{figure}

\subsection{Sub-Trajectory Dense Rewards}
\label{sec:dense_reward}

In MMPO, we evaluate the agent's performance through the lens of \textit{sub-trajectories}. A sub-trajectory $\tau_{\leq k}^{(i)}$ represents the reasoning path of the $i$-th sample from the initial step up to turn $k$ ($1 \leq k \leq T$). 

Unlike standard sparse-reward RL, we assign a reward $R_k^{(i)}$ to each sub-trajectory that incorporates both local memory quality and the global task outcome. Let $m_k^{(i)}$ be the memory summary at turn $k$, and $r_{\text{final}}^{(i)} \in [0, 1]$ be the terminal outcome reward of the $i$-th complete trajectory (e.g., token-level F1 score for QA tasks, following MemAgent~\citep{MemAgent} and MEM1~\citep{mem1}). A na\"ive linear formulation $-\alpha \cdot \mathcal{H}_{\text{BE}} + r_{\text{final}}$ directly couples the unbounded entropy with the bounded outcome reward, creating scale mismatch and numerical instability. To address this, we normalize the entropy signal via a sigmoid transformation:
\begin{equation}
    R_k^{(i)} = \alpha \cdot \underbrace{\sigma\!\Big(-\mathcal{H}_{\text{BE}}(m_k^{(i)})\Big)}_{\text{Normalized BE Reward} \in (0,1)} + \underbrace{r_{\text{final}}^{(i)}}_{\text{Outcome Reward} \in [0,1]}
\end{equation}
where $\sigma(\cdot)$ is the sigmoid function and $\alpha > 0$ controls the relative weight of the intrinsic belief reward against the outcome reward. Since $\mathcal{H}{\text{BE}}$ is non-negative and may vary in scale across tasks, the sigmoid transformation converts it into a bounded reward signal: lower Belief Entropy yields a larger reward, while higher Belief Entropy yields a smaller reward. Placing $\alpha$ outside the sigmoid separates reward normalization from loss weighting. By attaching $r{\text{final}}$ to every sub-trajectory within the same reasoning path, the reward still remains anchored to final task success, while the Belief Entropy term provides dense intermediate supervision for clearer memory states.

\subsection{Turn-Level Advantage Estimation}
\label{sec:grpo_advantage}

To optimize the policy without the instability of a learned value network, MMPO utilizes Group Relative Advantage Estimation. For each task, we sample a group of $N$ complete trajectories. At each reasoning depth $k$, we aggregate the sub-trajectory rewards across the group $\{R_k^{(1)}, R_k^{(2)}, \dots, R_k^{(G)}\}$. 
The \textit{sub-trajectory advantage} $\hat{A}_k^{(i)}$ is computed by standardizing the rewards within this group at turn $k$:
\begin{equation}
    \hat{A}_k^{(i)} = \frac{R_k^{(i)} - \text{mean}(R_k)}{\text{std}(R_k)}
\end{equation}
By comparing sub-trajectories at the same step, $\hat{A}_k^{(i)}$ isolates the relative contribution of the $i$-th path's memory policy to the task's progress. A positive advantage indicates that the $k$-step prefix of the $i$-th sample is superior to its peers in balancing memory quality and goal attainment.


To update the memory policy for generating $m_t$, we must aggregate the advantages of all future sub-trajectories influenced by $m_t$. Since the summary at turn $t$ serves as the context for all subsequent steps $k \in \{t, t+1, \dots, T\}$, its overall quality is reflected in the performance of all these overlapping sub-trajectories.

We compute the \textit{turn-level advantage} $A_t^{(i)}$ by averaging the advantages of all sub-trajectories containing turn $t$:
\begin{equation}
    A_t^{(i)} = \frac{1}{T - t + 1} \sum_{k=t}^T \hat{A}_k^{(i)}
\end{equation}
This aggregation mechanism provides a robust credit assignment: a memory summary $m_t$ is reinforced if it leads to a sequence of subsequent states that are consistently clearer and more successful than the alternatives in the group.

\subsection{Optimization Objective}
\label{sec:optimization_objective}

The turn-level advantage $A_t^{(i)}$ is distributed to all tokens comprising the summary $m_t$. Let the summary consist of $L$ tokens $\{w_1, \dots, w_L\}$. We optimize the memory policy $\pi_\theta$ using the clipped PPO surrogate objective:
\begin{equation}
    \mathcal{J}_{\text{MMPO}}(\theta) = \mathbb{E} \left[ \sum_{t=1}^T \sum_{j=1}^L \min \Big( \rho_{t, j}(\theta) A_t^{(i)}, \, \text{clip}(\rho_{t, j}(\theta), 1-\epsilon, 1+\epsilon) A_t^{(i)} \Big) \right] - \beta \mathbb{D}_{\text{KL}}(\pi_\theta \| \pi_{\text{ref}})
    \label{eq:mmpo_objective}
\end{equation}
where $\rho_{t, j}(\theta)$ is the token-level importance ratio and $\mathbb{D}_{\text{KL}}$ is the penalty against the reference model $\pi_{\text{ref}}$. This objective fine-tunes the model to generate summaries that minimize belief deviation and maximize the probability of task success across long horizons.

\subsection{Algorithm Overview}
\label{sec:algorithm}
The complete MMPO training procedure is given in Algorithm~\ref{alg:mmpo} (Appendix~\ref{app:algorithm}). At each iteration, the policy samples a group of trajectories, computes per-turn Belief Entropy, constructs sub-trajectory rewards, and updates parameters via the clipped objective in Eq.~\eqref{eq:mmpo_objective}.

\paragraph{Implementation Details.}
At each turn $t$, the memory policy receives the previous summary $m_{t-1}$ and the current observation $o_t$, and generates a new summary $m_t$. Belief Entropy is computed using Eq.~\ref{eq:belief_entropy_token}. The anchor question is adapted per task type: for QA tasks, ``Based on current memory, what is our task progress and what information is still needed?''; for agentic tasks with tool use, ``Based on current memory, what is our task progress and what steps remain?''

\begin{table}[t]
\centering
\caption{Main results on RULER-HotpotQA across context lengths.  MMPO follows the same recursive memory workflow as RL-MemAgent and adds Belief Entropy supervision during training.}
\label{tab:ruler_hqa}
{
\footnotesize
\renewcommand{\arraystretch}{1.12}
\setlength{\tabcolsep}{3pt}
\resizebox{\linewidth}{!}{
\begin{tabular}{lcccccccccc}
\toprule
\textbf{Model} 
& \textbf{7K} & \textbf{14K} & \textbf{28K} & \textbf{56K} 
& \textbf{112K} & \textbf{224K} & \textbf{448K} & \textbf{896K} 
& \textbf{1.75M} & \textbf{3.5M} \\
\midrule
QwenLong-L1-32B 
& 72.66 & 75.00 & 72.66 & 60.94 & 31.25 & 17.19 & 13.28 & 11.72 & N/A & N/A \\
Qwen2.5-14B-1M 
& 60.16 & 60.94 & 50.00 & 57.03 & 50.00 & 37.50 & 8.59 & 0.00 & N/A & N/A \\
Qwen2.5-7B-1M 
& 61.72 & 56.25 & 53.91 & 55.47 & 51.56 & 33.59 & 12.50 & 0.00 & N/A & N/A \\
\midrule
DS-Qwen-32B 
& 70.31 & 66.41 & 65.62 & 46.88 & 23.44 & 13.28 & 7.81 & 7.03 & N/A & N/A \\
DS-Qwen-14B 
& 64.06 & 64.84 & 57.03 & 40.62 & 14.84 & 8.59 & 3.12 & 6.25 & N/A & N/A \\
DS-Qwen-7B 
& 30.47 & 12.50 & 3.12 & 0.00 & 0.00 & 0.78 & 0.00 & 0.00 & N/A & N/A \\
\midrule
RL-MemAgent-14B 
& 80.47 & 82.03 & 82.03 & 83.59 & 81.25 & 77.34 & 79.69 & 75.78 & 78.91 & 71.09 \\

\rowcolor{MMPOGray}
\textbf{MMPO-14B} 
& 82.16 & \textbf{83.38} & \textbf{84.32} & \textbf{83.75} & 82.47 & \textbf{80.59} & \textbf{80.91} & \textbf{80.69} & \textbf{79.77} & \textbf{76.47} \\

RL-MemAgent-7B 
& 81.25 & 81.25 & 82.03 & 80.47 & 79.69 & 75.78 & 76.56 & 74.22 & 77.34 & 71.88 \\

\rowcolor{MMPOGray}
\textbf{MMPO-7B} 
& \textbf{82.38} & 82.81 & 81.03 & 82.98 & \textbf{82.81} & 79.56 & 78.12 & 79.69 & 78.91 & 75.22 \\
\bottomrule
\end{tabular}
}
}
\vspace{-0.5em}
\end{table}

\section{Experiments}
\label{sec:experiments}
\vspace{-0.5em}
We evaluate MMPO against two representative memory-agent frameworks: MemAgent\citep{MemAgent} and MEM1\citep{mem1}. The MemAgent comparison tests whether Belief Entropy supervision improves recursive memory summarization under extreme context-length scaling on RULER-HotpotQA. The MEM1 comparison tests whether the same supervision is complementary to MEM1's outcome-based training on multi-objective QA and WebShop. Since MMPO keeps the corresponding memory workflow unchanged and only augments the training signal, we focus on task performance under the original evaluation protocols.

\subsection{Experimental Setup}
For the \textbf{MemAgent}-based comparison, we evaluate MMPO on RULER-HotpotQA following the MemAgent setting~\citep{MemAgent}. RULER-HotpotQA combines HotpotQA-style multi-hop questions with controllable long-context scaling, requiring the agent to update memory over progressively longer distractor contexts. This benchmark tests whether a memory agent can retain task-relevant evidence under extreme context-length scaling. We train MMPO with the same recursive memory workflow, so that the comparison isolates the effect of adding Belief Entropy supervision to intermediate memory states. For the \textbf{MEM1}-based comparison, we evaluate MMPO under both multi-objective QA and WebShop~\citep{yao2022webshop}. In multi-objective QA, the agent is trained on 2-objective tasks and evaluated on longer objective horizons, requiring memory to track multiple unresolved information needs. In WebShop, the agent interacts with the environment through search and click actions, testing whether the same memory supervision benefits interactive decision-making beyond retrieval-based QA. We further compare with additional agent baselines, including A-MEM~\citep{amem}, Search-R1~\citep{searchr1}, and DeepResearcher~\citep{deepresearcher}. More details of the compared memory-agent frameworks are provided in Appendix~\ref{app:framework_details}.
\subsection{Main Results}
\paragraph{Comparison with MemAgent}
Table~\ref{tab:ruler_hqa} compares MMPO with MemAgent and other long-context/reasoning baselines on RULER-HotpotQA. We evaluate MMPO with both Qwen2.5-7B and Qwen2.5-14B backbones, using the same memory workflow as MemAgent. MMPO improves the average accuracy over RL-MemAgent for both model sizes. The improvement is especially clear in the long-context regime: from 224K to 3.5M context length, MMPO improves accuracy over RL-MemAgent by an average of $+3.14\%$ on Qwen2.5-7B and $+3.12\%$ on Qwen2.5-14B. The largest gains are $+5.47\%$ at 896K for Qwen2.5-7B and $+5.38\%$ at 3.5M for Qwen2.5-14B. These results suggest that Belief Entropy provides useful intermediate supervision for recursive memory summarization, helping the memory policy preserve clearer summary-induced beliefs and reduce noise accumulation over long contexts.

\paragraph{Comparison with MEM1}
Table~\ref{tab:mem1_comparison} evaluates whether MMPO is complementary to MEM1's outcome-based memory training. On multi-objective QA, MMPO improves over MEM1-QA across the evaluated objective horizons, with larger gains at harder settings such as 8-objective and 16-objective QA, where the agent must maintain multiple unresolved information needs over longer trajectories. On WebShop, MMPO also improves over MEM1-WebShop (see Table~\ref{tab:webshop_results}), showing that the proposed supervision is not limited to retrieval-based QA but also benefits interactive tool-use tasks. Since the underlying memory architecture is unchanged, these improvements suggest that the gain comes from the denser memory-quality signal introduced by Belief Entropy.

\begin{table}[t]
\centering
\caption{Comparison with MEM1 on Multi-objective QA. We follow the MEM1 evaluation protocol and report task scores. Following MEM1, EM/F1 are aggregated over objectives. }
\label{tab:mem1_comparison}
\small

\setlength{\tabcolsep}{0pt}
\begin{tabular*}{\linewidth}{@{\extracolsep{\fill}} lcccccc}
\toprule
\multirow{2}{*}{\textbf{Model}}
& \multicolumn{2}{c}{\textbf{2-Objective}}
& \multicolumn{2}{c}{\textbf{8-Objective}}
& \multicolumn{2}{c}{\textbf{16-Objective}} \\
\cmidrule(lr){2-3} \cmidrule(lr){4-5} \cmidrule(lr){6-7}
& \textbf{EM} $\uparrow$ & \textbf{F1} $\uparrow$
& \textbf{EM} $\uparrow$ & \textbf{F1} $\uparrow$
& \textbf{EM} $\uparrow$ & \textbf{F1} $\uparrow$ \\
\midrule
Qwen2.5-14B-Inst
& 0.732 & 0.902 & 1.55 & 1.87 & 0.567 & 0.703 \\
Qwen2.5-7B-Inst
& 0.268 & 0.366 & 0.87 & 1.10 & 0.165 & 0.213 \\
Qwen2.5-7B-Inst (A-MEM)
& 0.286 & 0.371 & 1.13 & 1.43 & 0.730 & 0.961 \\
Qwen2.5-7B-Inst (truncate)
& 0.262 & 0.336 & 0.97 & 1.23 & 0.396 & 0.497 \\
Search-R1
& 0.452 & 0.531 & 0.064 & 0.080 & 0.009 & 0.011 \\
DeepResearcher
& 0.536 & 0.650 & 0.73 & 0.90 & 0.071 & 0.106 \\
MEM1-QA
& 0.709 & 0.838 & 1.87 & 2.31 & 1.97 & 2.39 \\
\textbf{MMPO}
& \textbf{0.725} & \textbf{0.852}
& \textbf{2.15} & \textbf{2.63}
& \textbf{2.43} & \textbf{2.84} \\
\bottomrule
\end{tabular*}
\vspace{-1em}
\end{table}

\begin{table}[t]
\centering
\begin{minipage}[t]{0.43\linewidth}
\vspace{0pt}
\centering
\captionof{table}{WebShop results under the MEM1 evaluation protocol. }
\label{tab:webshop_results}
\small
\setlength{\tabcolsep}{6pt}
\begin{tabular*}{\linewidth}{@{\extracolsep{\fill}}lc}
\toprule
\textbf{Method} & \textbf{Avg. Reward} $\uparrow$ \\
\midrule
Qwen2.5-7B & 18.42 \\
Qwen2.5-14B & 12.34 \\
MEM1-WebShop & 70.87 \\
\textbf{MMPO} & \textbf{77.25} \\
\bottomrule
\end{tabular*}
\end{minipage}
\hfill
\begin{minipage}[t]{0.53\linewidth}
\vspace{0pt}
\centering
\captionof{table}{Anchor question ablation on Ruler HQA with Qwen2.5-7B at 56K context length.}
\label{tab:ablation_anchor}
\small
\setlength{\tabcolsep}{0pt}
\begin{tabular*}{\linewidth}{@{\extracolsep{\fill}}lcc}
\toprule
\textbf{Anchor Question} & \textbf{Acc.} $\uparrow$ & $\Delta\mathcal{H}_{\text{BE}}$ \textbf{Corr.} \\
\midrule
Outcome Only & 80.47 & -- \\
Direct-answer probe & 78.17 & $-0.54$ \\
Gap-only probe & 82.02 & $-0.62$ \\
\textbf{Progress + gap} & \textbf{82.98 } & $\mathbf{-0.68}$ \\
\bottomrule
\end{tabular*}
\end{minipage}
\end{table}

\subsection{Analysis}
\label{sec:ablation}
\paragraph{Anchor Question Ablation.}
Table~\ref{tab:ablation_anchor} compares anchor-question designs on RULER-HQA with Qwen2.5-7B at 56K context length. \textit{Outcome Only} is the standard RLVR baseline using only final task reward. The \textit{Direct-answer probe} asks for the final answer from memory, the \textit{Gap-only probe} asks for missing evidence, and our default \textit{Progress + gap} probe asks for both task progress and missing information. The direct-answer probe remains correlated with task accuracy ($r=-0.54$), but underperforms Outcome Only, suggesting that rewarding low answer entropy can encourage premature confidence before sufficient evidence is collected. In contrast, the gap-only and progress+gap probes better target intermediate memory quality by exposing unresolved information needs. The progress+gap probe performs best, showing that jointly tracking progress and missing information provides a more informative memory-quality signal.

\begin{figure}[t]
    \centering
    \includegraphics[width=\textwidth]{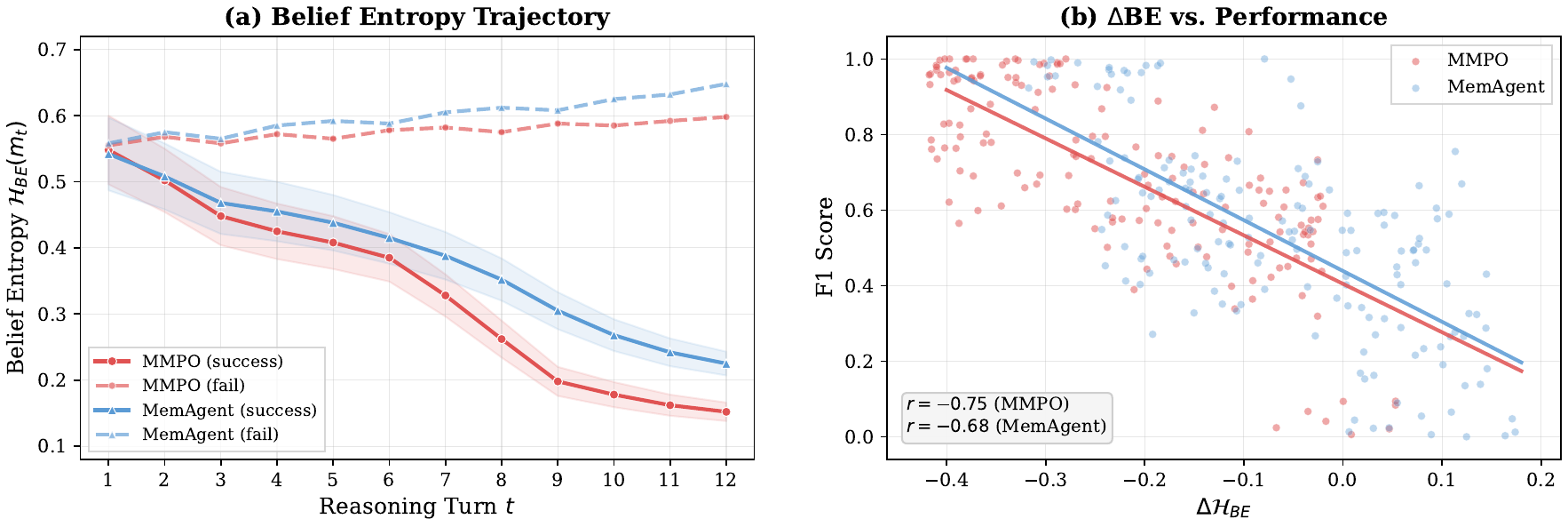}
    \caption{\textbf{Belief Entropy analysis.} 
    \textbf{(a)} Belief Entropy trajectories over reasoning turns at 56K context length. Successful trajectories show consistent entropy decrease, while failed trajectories stagnate or increase. 
    \textbf{(b)} Correlation between total entropy reduction $\Delta\mathcal{H}_{\text{BE}}$ and task accuracy across 500 test episodes. MMPO strengthens this correlation compared with MemAgent, supporting Belief Entropy as a proxy for intermediate memory quality.}
    \label{fig:analysis}
    \vspace{-1em}
\end{figure}

\paragraph{Belief Entropy Dynamics.}
Figure~\ref{fig:analysis} analyzes how Belief Entropy evolves during long-horizon reasoning. Successful trajectories generally show decreasing $\mathcal{H}_{\mathrm{BE}}$ as evidence accumulates, while failed trajectories tend to stagnate or increase. MMPO also strengthens the correlation between entropy reduction and task accuracy, suggesting that Belief Entropy captures useful intermediate information about memory quality. Figure~\ref{fig:analysis} analyzes Belief Entropy dynamics during long-horizon reasoning. Successful trajectories show decreasing entropy as evidence accumulates, while failed trajectories stagnate or increase. MMPO yields a steeper entropy decline than MemAgent and strengthens the correlation between entropy reduction and task accuracy, indicating that BE supervision better aligns intermediate memory clarity with final task success.

Additional analyses, including comparison with alternative proxy signals and computational overhead, are provided in Appendix~\ref{app:more_results}.
\section{Related Work}
\paragraph{Memory-Augmented LLM Agents.}
Managing long interaction histories is a central challenge for long-horizon LLM agents. Early methods rely on truncation or retrieval-augmented context construction~\citep{long-context-survey}, while recent memory agents compress trajectories into compact memory states. MemGPT~\citep{MemGPT} introduces an operating-system-inspired memory hierarchy, and Mem0~\citep{Mem0} and MemOS~\citep{MemOS} develop learnable memory management layers. More recent works further formulate memory as a trainable policy within the agent loop~\citep{du2026memory, MemAgent, mem1}. However, they are typically optimized by final task outcomes, leaving intermediate summaries weakly supervised. MMPO addresses this gap by supervising the quality of summary-induced beliefs during memory optimization.
\vspace{-0.5em}
\paragraph{Reinforcement Learning for LLMs.}
RLHF and related outcome-based objectives are widely used for LLM alignment~\citep{Deepseek-R1}. To improve credit assignment in multi-step reasoning, process-supervision methods such as PRM~\citep{PRM8K}, PRIME~\citep{PRIME}, and Miracle~\citep{Miracle} assign rewards to intermediate reasoning states, while GRPO~\citep{Deepseek-R1} stabilizes RL through group-relative normalization without a value model. In memory-agent settings, RL4LRM~\citep{RL4LRM}, MemAlpha~\citep{MemAlpha}, MemAgent~\citep{MemAgent}, and MEM1~\citep{mem1} apply RL to memory policies, but still largely rely on outcome-level rewards. MMPO instead defines dense rewards at intermediate memory states, targeting summary-induced belief uncertainty rather than generic reasoning quality.
\vspace{-0.5em}
\paragraph{Belief States and Uncertainty Estimation.}
Belief states are central to decision-making under partial observability~\citep{astrom1965optimal, smallwood1973optimal, kaelbling1998planning}, with related work studying history compression through predictive state representations and information-theoretic sufficient statistics~\citep{littman2001predictive, still2012information}. Recent studies identify belief deviation as a key failure mode of long-horizon LLM agents~\citep{zoureducing}, motivating memory mechanisms that preserve reliable state estimates. In parallel, predictive entropy, semantic entropy, verbalized confidence, and self-consistency have been used to quantify LLM uncertainty~\citep{kadavath2022language, kuhn2023semantic}. MMPO connects these directions by using a self-supervised entropy signal as dense supervision for memory-policy optimization, rather than for selective prediction.

\section{Conclusion}
We introduced Metacognitive Memory Policy Optimization (MMPO), a memory optimization framework for long-horizon LLM agents. MMPO uses Belief Entropy to estimate the uncertainty of the summary-induced belief and provides dense supervision for intermediate memory states. This allows the memory policy to optimize not only final task success, but also the reliability of the evolving memory. Experiments on long-horizon agent tasks show that MMPO consistently improves over outcome-based memory RL baselines.

\bibliographystyle{plainnat}
\bibliography{ref}


\appendix

\section{MMPO Algorithm}
\label{app:algorithm}

\begin{algorithm}[h]
\caption{Metacognitive Memory Policy Optimization (MMPO)}
\label{alg:mmpo}
\begin{algorithmic}[1]
\REQUIRE Policy $\pi_\theta$, reference policy $\pi_{\text{ref}}$, anchor question $q$, group size $N$, max turns $T$, coefficients $\alpha, \beta, \epsilon$
\FOR{each training iteration}
    \STATE Sample a batch of tasks from the training set
    \FOR{each task}
        \STATE Sample $N$ complete trajectories $\{\tau^{(1)}, \dots, \tau^{(N)}\}$ using $\pi_\theta$
        \FOR{each trajectory $\tau^{(i)}$, each turn $t = 1, \dots, T$}
            \STATE Generate memory summary: $m_t^{(i)} \sim \pi_\theta(\cdot | m_{t-1}^{(i)}, o_t)$
            \STATE Compute $\mathcal{H}_{\text{BE}}(m_t^{(i)})$ via token-level predictive entropy following Eq.~\ref{eq:belief_entropy_token}
        \ENDFOR
        \STATE Obtain terminal outcome reward $r_{\text{final}}^{(i)} \in [0,1]$ for each trajectory
        \FOR{each depth $k = 1, \dots, T$}
            \STATE Compute sub-trajectory reward: $R_k^{(i)} = \alpha \cdot \sigma\!\big(-\mathcal{H}_{\text{BE}}(m_k^{(i)})\big) + r_{\text{final}}^{(i)}$
            \STATE Compute group-relative advantage: $\hat{A}_k^{(i)} = (R_k^{(i)} - \text{mean}(R_k)) / \text{std}(R_k)$
        \ENDFOR
        \FOR{each turn $t = 1, \dots, T$}
            \STATE Aggregate turn-level advantage: $A_t^{(i)} = \frac{1}{T-t+1} \sum_{k=t}^{T} \hat{A}_k^{(i)}$
        \ENDFOR
    \ENDFOR
    \STATE Update $\theta$ via clipped PPO objective $\mathcal{J}_{\text{MMPO}}(\theta)$ (Eq.~6)
\ENDFOR
\end{algorithmic}
\end{algorithm}

\section{Summary-Induced Belief: Architectural Justification}
\label{app:summary_belief}

This appendix justifies why the belief of a summary-based memory agent is conditioned on the textual memory $m_t$ rather than on the full interaction history $h_t$. The key point is architectural: once the history is compressed into memory, downstream reasoning and action selection can only access the information preserved in $m_t$.

\paragraph{From Full-History Belief to Summary-Induced Belief.}
In a standard POMDP, the full-history belief is
\begin{equation}
    b_t(s) \triangleq P(s_t=s \mid h_t),
\end{equation}
where $h_t=\{o_{\leq t},a_{<t}\}$ denotes the full interaction history. In a summary-based agent, however, the memory policy compresses this history into a bounded textual memory,
\begin{equation}
    m_t \sim \pi_{\mathrm{mem}}(\cdot \mid h_t),
\end{equation}
and the action policy subsequently conditions on $m_t$ rather than on $h_t$:
\begin{equation}
    a_t \sim \pi_{\mathrm{act}}(\cdot \mid m_t).
\end{equation}
Therefore, the belief used by the agent is induced by the compressed memory:
\begin{equation}
    b^M_t(s) \triangleq P(s_t=s \mid m_t).
\end{equation}
This is not a claim that $m_t$ is a sufficient statistic of $h_t$; rather, it is the belief imposed by the summary-based architecture.

\paragraph{Information Constraint Induced by Summarization.}
Because $m_t$ is generated from $h_t$, the variables form the Markov chain
\begin{equation}
    s_t \rightarrow h_t \rightarrow m_t.
\end{equation}
By the data processing inequality~\citep{cover2006elements},
\begin{equation}
    I(s_t;m_t) \leq I(s_t;h_t).
\end{equation}
Thus, summarization cannot increase the information about the latent task state. Any state information not preserved in $m_t$ is unavailable to the action policy, since downstream reasoning no longer conditions on the original history.

\paragraph{Mixture View of Summary-Induced Belief.}
The summary-induced belief can also be viewed as a mixture over full-history beliefs compatible with the same memory. Using the Markov property $s_t \perp m_t \mid h_t$, we have
\begin{equation}
    P(s_t \mid m_t)
    =
    \sum_{h_t}
    P(s_t \mid h_t,m_t)P(h_t\mid m_t)
    =
    \sum_{h_t}
    P(s_t \mid h_t)P(h_t\mid m_t).
\end{equation}
Therefore, $b_t^M$ aggregates the full-history beliefs of all histories that could have produced the same summary $m_t$. If different histories requiring different decisions are compressed into similar or ambiguous summaries, their state estimates become mixed under $P(s_t\mid m_t)$. This explains why semantic noise, omitted evidence, or conflated entities in recursive summaries can induce belief deviation in downstream reasoning.

\section{Information-Theoretic Justification for Belief Entropy}
\label{app:be_proof}

We provide an information-theoretic justification for using Belief Entropy as an intermediate signal for memory optimization. The ideal objective in Eq.~\ref{eq:ideal_objective} minimizes the conditional uncertainty $H(s_t \mid m_t)$ of the latent task state given the current memory. Since $s_t$ is not directly observable in open-ended LLM-agent settings, Belief Entropy uses a state-probing anchor question to expose part of this uncertainty through the model's response distribution.

\paragraph{Anchor Response as a State-Probing Signal.}
Let $q$ denote the anchor question and let $y$ denote the model's response conditioned on the current memory $m_t$. The anchor question is designed to make the response depend on task-state information preserved in memory: the progress component probes the agent's current estimate of the task state, while the information-gap component probes uncertainty that remains unresolved. Thus, $y$ is not treated as an arbitrary model output; it is a response whose uncertainty is intended to reflect how clearly $m_t$ specifies the current task state.

Belief Entropy measures this response uncertainty:
\begin{equation}
    \mathcal{H}_{\mathrm{BE}}(m_t)
    =
    H(y \mid m_t,q).
\end{equation}
If $m_t$ preserves the task-relevant information needed to answer the anchor question, the response distribution should be more concentrated. If $m_t$ omits key evidence or contains semantic noise, the model must resolve more uncertainty when answering the anchor question, leading to higher response entropy.

\paragraph{Chain-Rule Decomposition.}
For the joint distribution over $(s_t,m_t,q,y)$, the conditional entropy decomposes as
\begin{equation}
    H(y \mid m_t,q)
    =
    H(y \mid m_t,q,s_t)
    +
    I(y; s_t \mid m_t,q).
    \label{eq:be_decomposition}
\end{equation}
The first term captures state-conditioned response uncertainty, such as verbalization variability and generation noise after the latent state is specified. The second term captures residual state uncertainty exposed through the anchor response: if the current memory already resolves the task state relevant to $q$, the anchor response carries less additional dependence on $s_t$.

This residual term is not identical to the full conditional entropy $H(s_t \mid m_t)$; it is the part of state uncertainty that is visible through the anchor response. Nevertheless, it is controlled by the remaining uncertainty about $s_t$ under the current memory:
\begin{equation}
    I(y; s_t \mid m_t,q)
    \leq
    H(s_t \mid m_t,q).
\end{equation}
Since $q$ is fixed by design when comparing memories at the same turn, reducing the uncertainty of the summary-induced belief $P(s_t\mid m_t)$ also reduces the amount of state uncertainty that can be exposed through the anchor response.

\paragraph{Assumptions and Implication.}
The proxy relies on two conditions.

\textbf{A1: Relevance.}
The anchor response should be relevant to the underlying task state. Formally, the response should carry nontrivial information about $s_t$ under the anchor question:
\begin{equation}
    I(y; s_t \mid q) > 0.
\end{equation}
This does not require the response to identify the full latent state; it only requires that the anchor question probes task-relevant aspects such as current progress, missing evidence, satisfied constraints, or remaining actions.

\textbf{A2: Memory Grounding.}
Response uncertainty should be primarily governed by the task-state information preserved in $m_t$. Consider two memories $m_t$ and $m_t^+$, where $m_t^+$ induces a more reliable estimate of the latent task state:
\begin{equation}
    H(s_t \mid m_t^+) \leq H(s_t \mid m_t).
\end{equation}
Under a relevant and memory-grounded anchor question, this improvement reduces the residual dependence of the anchor response on the latent state:
\begin{equation}
    I(y; s_t \mid m_t^+,q)
    \leq
    I(y; s_t \mid m_t,q).
\end{equation}
If the state-conditioned response uncertainty is approximately stable across the two memories,
\begin{equation}
    H(y \mid m_t^+,q,s_t)
    \approx
    H(y \mid m_t,q,s_t),
\end{equation}
then Eq.~\ref{eq:be_decomposition} gives
\begin{equation}
    \mathcal{H}_{\mathrm{BE}}(m_t^+)
    \lesssim
    \mathcal{H}_{\mathrm{BE}}(m_t).
\end{equation}
Therefore, Belief Entropy serves as an anchor-probed proxy for the belief-preservation objective in Eq.~\ref{eq:ideal_objective}: lower response uncertainty indicates that the current memory more reliably resolves the task-state information exposed by the anchor question.

Finally, MMPO does not use Belief Entropy as a standalone correctness reward. The entropy signal provides dense intermediate credit for memory clarity, while the verifiable outcome reward anchors optimization to final task success. This design reduces the risk that the policy optimizes only for response confidence rather than useful memory content.

\section{Details of Compared Memory-Agent Frameworks}
\label{app:framework_details}
This appendix summarizes the two memory-agent frameworks used in our experiments and clarifies how MMPO is applied to them. 

\paragraph{MemAgent Framework.}
MemAgent~\citep{MemAgent} formulates long-context reasoning as a recurrent memory-update process. At each turn, the model receives the task query, the previous memory, and the current context segment, and produces an updated memory summary. After processing all segments, the final answer is generated from the accumulated memory. This design enables a fixed context window to process inputs much longer than the model's native context length, but the memory policy is mainly optimized through final task outcomes. In our MemAgent-based experiments, MMPO keeps this recurrent memory workflow unchanged and adds Belief Entropy supervision to the intermediate memory summaries. Thus, the comparison evaluates whether dense belief-quality feedback improves recursive memory summarization beyond outcome-only memory RL.

\paragraph{MEM1 Framework.}
MEM1~\citep{mem1} studies long-horizon agents that jointly maintain internal memory and perform task-directed reasoning. Instead of relying only on raw interaction history, the agent maintains a compact internal memory state across steps and uses it to support subsequent reasoning, querying, or environment interaction. This framework is evaluated in both multi-objective QA and WebShop-style interactive tasks, where the agent must preserve multiple information needs or action-relevant constraints over a long trajectory. In our MEM1-based experiments, MMPO keeps the memory workflow and task interaction format unchanged, while adding Belief Entropy as an intermediate reward for memory states. This tests whether the proposed supervision can improve memory optimization not only in recursive summarization, but also in broader memory-augmented agent workflows.

\section{More Results}
\label{app:more_results}

\subsection{Full Anchor Question Robustness Study}

Table~\ref{tab:ablation_anchor_full} reports the anchor-question robustness study across all evaluated RULER-HotpotQA context lengths with Qwen2.5-7B. We compare three anchor designs: a direct-answer probe, a gap-only probe, and the default progress+gap probe. The direct-answer probe underperforms Outcome Only on average, suggesting that directly rewarding answer confidence can encourage premature certainty when the memory is still incomplete. In contrast, the gap-only and progress+gap probes better target intermediate memory quality by exposing unresolved information needs. The progress+gap probe achieves the best average performance, indicating that jointly tracking task progress and missing information provides a more informative signal for memory optimization.

\begin{table}[h]
\centering
\caption{Full anchor question robustness study on RULER-HotpotQA with Qwen2.5-7B. All values are accuracy (\%). Outcome Only denotes the standard RLVR setting without Belief Entropy supervision.}
\label{tab:ablation_anchor_full}
\scriptsize
\setlength{\tabcolsep}{0pt}
\begin{tabular*}{\linewidth}{@{\extracolsep{\fill}} lccccccccc}
\toprule
\textbf{Anchor Question} 
& \textbf{7K} & \textbf{14K} & \textbf{28K} & \textbf{56K} 
& \textbf{112K} & \textbf{224K} & \textbf{448K} & \textbf{896K} 
& \textbf{Avg.} \\
\midrule
Outcome Only 
& 81.25 & 81.25 & 82.03 & 80.47 & 79.69 & 75.78 & 76.56 & 74.22 & 78.91 \\
Direct-answer probe 
& 80.10 & 79.80 & 79.20 & 78.17 & 78.60 & 75.10 & 75.90 & 74.80 & 77.71 \\
Gap-only probe 
& 81.90 & 82.10 & 80.80 & 82.02 & 82.20 & 78.70 & 77.30 & 78.40 & 80.43 \\
\textbf{Progress + gap}
& \textbf{82.38} & \textbf{82.81} & 81.03 & \textbf{82.98} 
& \textbf{82.81} & \textbf{79.56} & \textbf{78.12} & \textbf{79.69} & \textbf{81.17} \\
\bottomrule
\end{tabular*}
\end{table}

We use the following anchor prompts. The direct-answer probe asks: ``Based on current memory, what is the answer to the question?'' The gap-only probe asks: ``Based on current memory, what key information is still needed to answer the question?'' The progress+gap probe asks: ``Based on current memory, what is our task progress and what information is still needed?'' The stronger performance of the progress+gap probe suggests that Belief Entropy benefits from explicitly tracking both the current task state and the remaining uncertainty.

\subsection{Comparison with Alternative Proxy Signals}

To examine whether Belief Entropy captures task-relevant memory quality rather than generic model confidence, we compare it with several low-cost proxy signals. As shown in Table~\ref{tab:proxy_comparison}, Belief Entropy achieves the strongest correlation with task accuracy and yields the best downstream performance when used as the dense reward signal.

\begin{table}[h]
\centering
\caption{Comparison of proxy signals as dense rewards on RULER-HotpotQA with Qwen2.5-7B at 56K context length. $|r|$ denotes the absolute Pearson correlation with task accuracy.}
\label{tab:proxy_comparison}
\small
\setlength{\tabcolsep}{8pt}
\begin{tabular}{lcc}
\toprule
\textbf{Proxy Signal} & $|r|$ \textbf{with Accuracy} & \textbf{Accuracy (\%)} \\
\midrule
Selected-token NLL (no anchor $q$) & 0.41 & 79.12 \\
Direct-answer entropy & 0.54 & 78.17 \\
Memory length & 0.22 & 79.84 \\
Random-question entropy & 0.31 & 80.05 \\
\midrule
\textbf{Belief Entropy ($\widehat{\mathcal{H}}_{\mathrm{BE}}$)} & \textbf{0.68} & \textbf{82.98} \\
\bottomrule
\end{tabular}
\end{table}

Selected-token NLL without the anchor question serves as a generic confidence baseline, but it does not directly probe task-state uncertainty. Direct-answer entropy remains correlated with task accuracy, but it can encourage premature answer confidence before sufficient evidence is collected. Memory length and random-question entropy show weaker correlations. These results indicate that the anchor-question-based Belief Entropy is better aligned with intermediate memory quality than generic confidence or length-based heuristics.

\subsection{Computational Overhead}

MMPO adds one additional forward pass per turn for Belief Entropy computation through the anchor-question response. On Qwen2.5-7B, this introduces approximately $12\%$ wall-clock overhead during training. At inference time, Belief Entropy is not required for standard decoding, so the additional cost can be removed; when used as an optional confidence signal, it introduces approximately $5\%$ overhead. Peak GPU memory remains unchanged because the anchor-question pass reuses the same model.

\section{Implementation Details}
\label{app:details}

\paragraph{Memory Generation.}
We follow the MemAgent recurrent memory workflow and prompt format. At each turn $t$, the model receives the task query, the previous memory $m_{t-1}$, and the current document chunk $o_t$, and generates an updated memory summary $m_t$. Following MemAgent, the context budget is allocated to a 1,024-token query, a 5,000-token document chunk, a 1,024-token memory, and a 1,024-token memory output, with the remaining tokens reserved for the chat template. Memory generation and final answer generation share the same model weights; MMPO keeps this workflow unchanged and only adds Belief Entropy supervision during training.

\paragraph{Training Setup.}
We train MMPO with group-relative policy optimization. We use AdamW with learning rate $1{\times}10^{-6}$ and a constant learning-rate schedule with linear warm-up. The KL coefficient is set to $1{\times}10^{-3}$, and the entropy loss is disabled. The group size is set to $G=16$. Each rollout batch contains $128$ trajectories, corresponding to $8$ prompts with $16$ sampled trajectories per prompt. The Belief Entropy reward weight is set to $\alpha=0.5$.

\paragraph{Prompt Templates.}
The memory update prompt follows MemAgent's original format. At each turn, the model receives the task problem, the previous memory, and the current article section, and then updates the memory by retaining previous relevant details while incorporating new useful information:
\begin{promptbox}
\small
\ttfamily
You are presented with a problem, a section of an article that may contain the answer to the problem, and a previous memory. Please read the provided section carefully and update the memory with the new information that helps to answer the problem. Be sure to retain all relevant details from the previous memory while adding any new, useful information.

\vspace{0.4em}
\noindent\textless problem\textgreater\\
\{prompt\}\\
\textless /problem\textgreater

\vspace{0.4em}
\noindent\textless memory\textgreater\\
\{memory\}\\
\textless /memory\textgreater

\vspace{0.4em}
\noindent\textless section\textgreater\\
\{chunk\}\\
\textless /section\textgreater

\vspace{0.4em}
\noindent Updated memory:
\end{promptbox}

For QA tasks, Belief Entropy is computed using the following anchor-question template:
\begin{promptbox}
\small
\ttfamily
Based on the problem and current memory, what is the current task progress and what information is still needed?

\vspace{0.4em}
\noindent\textless problem\textgreater\\
\{prompt\}\\
\textless /problem\textgreater

\vspace{0.4em}
\noindent\textless memory\textgreater\\
\{memory\}\\
\textless /memory\textgreater

\vspace{0.4em}
\noindent Your assessment:
\end{promptbox}

\section{Limitations}
\label{app:limitations}

Belief Entropy is designed as a practical proxy for summary-induced belief uncertainty, rather than a direct measurement of the latent task-state uncertainty. Its effectiveness therefore depends on whether the anchor question captures task-relevant uncertainty and whether the model's response uncertainty reflects the information preserved in memory. In this work, we mitigate this issue by using a task-state anchor question and combining the Belief Entropy signal with verifiable outcome rewards. Designing more adaptive or task-specific probes remains an interesting direction for future work.

\section{Impact Statement}
\label{app:impact}

This work aims to improve the reliability of long-horizon LLM agents by providing denser supervision for intermediate memory states. Better memory optimization may benefit applications that require sustained context tracking, such as long-document reasoning, multi-step question answering, interactive search, and task-oriented assistants.

However, MMPO does not guarantee factual correctness or safe autonomous behavior. Inaccurate memories, overconfident summaries, or poorly designed anchor questions may still lead to incorrect decisions, especially in long interactions or high-stakes settings. In addition, memory-based agents may process sensitive user information, so practical deployments should include appropriate privacy controls, retention policies, and user-facing transparency.



\end{document}